\documentclass[a4paper]{article}
\usepackage[english]{babel}
\usepackage{microtype,etex,listings,color,parskip}
\usepackage[margin=3cm]{geometry}
\usepackage{hyperref}
\usepackage{amsmath,amssymb, amsthm}
\usepackage{mathtools}
\usepackage{graphicx}
\usepackage{xcolor}
\usepackage[numbers]{natbib}
\usepackage{subcaption}
\usepackage{bm}
\usepackage{url}
\usepackage{endnotes}
\usepackage{titlecaps}
\usepackage{booktabs}

\usepackage{float}
\usepackage{tikz, pgfplots}
\usetikzlibrary{decorations.pathmorphing}
\tikzstyle{mybox} = [draw=black, fill=white, very thick,
    rectangle, rounded corners, inner sep=10pt, inner ysep=10pt]
\tikzstyle{fancytitle} =[fill=black, text=white, font=\bfseries]
\usepackage{algorithm}
\usepackage[noend]{algpseudocode}
\usepackage{blindtext}
\usepackage{titlesec}

\usepackage{booktabs}

\hypersetup{
    colorlinks=true,
    linkcolor=blue,
    filecolor=blue,      
    urlcolor=cyan,
    citecolor=blue,
}

\newcommand*{\NewTheorem}[1]{%
    \expandafter\providecommand\csname#1autorefname\endcsname{\MakeUppercase#1}%
    \newaliascnt{#1}{equation}%
    \newtheorem{#1}[#1]{\MakeUppercase#1}%
    \aliascntresetthe{#1}
}
\RequirePackage[T1]{fontenc}
\RequirePackage[utf8]{inputenc}

\RequirePackage{aliascnt}
\RequirePackage{amsmath, amssymb, amsthm}
\RequirePackage{mathtools}
\RequirePackage{microtype}
\RequirePackage{parskip}
\newtheoremstyle{hw-plain}{}{}{\itshape}{}{\bfseries}{ --- }{0pt}{}
\newtheoremstyle{hw-definition}{}{}{}{}{\bfseries}{ --- }{0pt}{}
\theoremstyle{hw-plain}
\NewTheorem{lemma}
\NewTheorem{theorem}

\theoremstyle{hw-definition}
\NewTheorem{definition}

\begin{document}

\title{Evaluating Uncertainty in Deep Gaussian Processes}
\author{Matthijs van der Lende \\ Jeremias Lino Ferrao \\ Niclas Müller-Hof \\ Bernoulli Institute, University of Groningen}
\date{} 
\maketitle

\begin{abstract}
Reliable uncertainty estimates are crucial in modern machine learning. Deep Gaussian Processes (DGPs) and Deep Sigma Point Processes (DSPPs) extend GPs hierarchically, offering promising methods for uncertainty quantification grounded in Bayesian principles. However, their empirical calibration and robustness under distribution shift relative to baselines like Deep Ensembles remain understudied. This work evaluates these models on regression (CASP dataset) and classification (ESR dataset) tasks, assessing predictive performance (MAE, Accuracy), calibration using Negative Log-Likelihood (NLL) and Expected Calibration Error (ECE), alongside robustness under various synthetic feature-level distribution shifts. Results indicate DSPPs provide strong in-distribution calibration leveraging their sigma point approximations. However, compared to Deep Ensembles, which demonstrated superior robustness in both performance and calibration under the tested shifts, the GP-based methods showed vulnerabilities, exhibiting particular sensitivity in the observed metrics. Our findings underscore ensembles as a robust baseline, suggesting that while deep GP methods offer good in-distribution calibration, their practical robustness under distribution shift requires careful evaluation. To facilitate reproducibility, we make our code available at \url{https://github.com/matthjs/xai-gp}.
\end{abstract}

\section{Introduction}
Modern machine learning models, particularly in high-stakes domains like healthcare and autonomous systems, require not only accurate predictions but also reliable estimates of uncertainty. While deep learning models excel at prediction tasks, their deterministic formulation leads to overconfident/uncalibrated uncertainty estimates, especially on out-of-distribution data. Gaussian processes (GPs) \cite{rasmussen_gaussian_2004, pml2Book} are nonparametric Bayesian models that provide well-calibrated predictive distributions, effectively capturing epistemic uncertainty—the uncertainty that arises from not knowing the true underlying model. They are used, for instance, in Bayesian optimization, where they act as a probabilistic surrogate model that is used to find promising hyperparameters in a hyperparameter space. There is also an interesting theoretical relationship between neural networks and GPs, namely, as the width of a neural network approaches infinity, the model converges to a GP with a specific kernel determined by their architecture \cite{pml2Book}.

Vanilla GPs rely on a kernel or covariance function to define a notion of similarity between datapoints. As a result, the function class that can be modeled is limited by the choice of kernel. Deep GPs (DGPs), introduced as multi-layer compositions of GPs \cite{damianou_deep_2013}, inherit the nonparametric flexibility of GPs while enabling richer hierarchical feature learning. More recently, Deep Sigma Point Processes (DSPPs) \cite{jankowiak2020deep} have been introduced as an alternative formulation of DGPs that admit a simpler training procedure.

Both models theoretically provide well-calibrated uncertainty quantification by propagating uncertainty through layers, yet their empirical calibration performance relative to uncertainty quantification baselines, such as deep ensembles, remains understudied. In particular, using metrics other than the negative log likelihood (NLL), which, although a proper scoring rule, can over-emphasize tail probabilities \cite{quinonero2005evaluating}. Additionally, there is a lack of empirical results on the use of non-Gaussian likelihoods with DGPs/DSPPs that arise in classification \cite{jankowiak2020deep}.

In particular, we are interested in determining how well calibrated the uncertainty quantification is of DGPs and DSPPs compared to a baseline (e.g., ensembles) for both regression and classification. For this purpose, we include several standardized uncertainty quantification evaluation methods, like expected calibration error and calibration plots. 
We will also look at out-of-distribution detection, in particular looking at how calibration behaves under distribution shift, which has not been studied in the model's respective papers, following the methodology from \citet{ovadia2019trustmodelsuncertaintyevaluating}.

\section{Background}\label{sec:theory} 


\subsection{Gaussian Processes}
A GP is a collection of random variables (RVs) \(\{f_{\text{GP}}(\mathbf{x}) \mid \mathbf{x} \in \mathcal{X}\}\), any finite number of which has a joint Gaussian distribution  \cite{rasmussen_gaussian_2004}. The index set \(\mathcal{X}\) is related to the input of some function \(f: \mathcal{X} \rightarrow \mathcal{Y}\) that we want to approximate. It can be interpreted as: At each point \(\mathbf{x} \in \mathcal{X}\), the output of a GP model is a RV denoted \(f_{\text{GP}}(\mathbf{x})\).

A GP, denoted \(f_{\text{GP}}(\mathbf{x}) \sim \mathcal{GP}(m(\mathbf{x}), k(\mathbf{x}, \mathbf{x}'))\), is fully specified by a mean function \(m : \mathcal{X} \rightarrow \mathbb{R}\) and covariance or kernel function     \(k : \mathcal{X} \times \mathcal{X} \rightarrow \mathbb{R}\) which are defined as:
\begin{equation}
    \begin{aligned}
    m(\mathbf{x}) &= \mathbb{E}[f_{\text{GP}}(\mathbf{x})], \\
    k(\mathbf{x}, \mathbf{x}') &= \mathbb{E}[(f_{\text{GP}}(\mathbf{x}) - m(\mathbf{x})) (f_{\text{GP}}(\mathbf{x}') -m(\mathbf{x}'))].
    \end{aligned}
\end{equation}

A GP offers a Bayesian approach to nonparametric regression or classification.  Without any data, the kernel function represents our prior belief about the function we are trying to model, as it encodes similarity between data points, with closer points having higher covariance. A GP can then be conditioned on a dataset, \(\mathcal{D}\), to get a posterior GP
 \(f_{\text{GP}}(\mathbf{x})\sim \mathcal{GP}(m_{\text{post}}(\mathbf{x}), k_{\text{post}}(\mathbf{x}, \mathbf{x}'))\),
 which is our posterior belief about the function. A common choice of kernel function is the Radial Basis Function (RBF) kernel: 
\begin{equation}
\label{eq:RBF}
k_{\text{RBF}}(\mathbf{x}, \mathbf{x}' ; l, \sigma_f) =  \sigma_f \exp\bigg(-\frac{\parallel \mathbf{x} - \mathbf{x}' \parallel^2}{2l^2} \bigg),
\end{equation}
with lengthscale and outputscale parameters \(l, \sigma_f \in \mathbb{R}\), which when used result in smooth infinitely differentiable functions being sampled from the GP.

Aside from computing the posterior, GP hyperparameters \(\theta\), such as parameters of the kernel, can be fit to data by maximizing the marginal log likelihood (MLL):
\begin{equation}
    p(\mathbf{y}|\mathbf{X}, \theta) = \int_{\mathbb{R}^N} p(\mathbf{y} | \mathbf{f}, \mathbf{X}) p(\mathbf{f}|\mathbf{X}, \theta) \, d\mathbf{f},
\end{equation}
where \(\mathbf{X} = (\mathbf{x}_1, \hdots \mathbf{x}_N) \in \mathbb{R}^{N \times D}\) are data points with targets \(\mathbf{y} \in \mathbb{R}^N\) where \(y_i = f(\mathbf{x}_i) + \epsilon_y, \epsilon_y \sim \mathcal{N}(0, \sigma_y)\) and \(\mathbf{f} \in \mathbb{R}^N\) are the instantiations of the function latent variables \(f_{\text{GP}}(\mathbf{x}_i)\) for each \(\mathbf{x}_i\). Both the MLL \(p(\mathbf{y}|\mathbf{X}, \theta)\) and the posterior distribution \(p(\mathbf{f}_*|\mathbf{X}, \mathbf{y}, \mathbf{X}_*)\) for test points \(\mathbf{X}_*\) can be computed in closed form, but this required an inversion of the kernel matrix over all data points, which is \(O(N^3)\) in time and \(O(N^2)\) in space.

\subsection{Sparse Variational Gaussian Processes}\label{sec:svgp}
Sparse Variational Gaussian Processes (SVGPs) \cite{titsias2009variational, hensman2014scalable} approximate the GP predictive posterior distribution through variational inference\footnote{See Appendix \ref{sec:vi} for a quick overview of variational inference.}.  This is done by introducing inducing variables \((f_{\text{GP}}(\mathbf{z}_1), \hdots, f_{\text{GP}}(\mathbf{z}_M))^\top = \mathbf{u}\) that depend on variational parameters \(\mathbf{Z} = (\mathbf{z}_1, \hdots, \mathbf{z}_M)\) called inducing points where \(M \ll N\). The inducing points together with the associated inducing variables serve as an approximation of the full dataset \(\mathcal{D} = (\mathbf{x}_i, y_i)_{i=1, ..., N}\)that the GP conditions on.

SVGPs consider the following variational distribution:
\begin{equation}
    q(\mathbf{f}, \mathbf{u}) = p(\mathbf{f}|\mathbf{u}, \mathbf{X}, \mathbf{Z}) q(\mathbf{u}) \quad q(\mathbf{u}) = \mathcal{N}(\mathbf{m}, \mathbf{S}).
    \label{eq:svgp}
\end{equation}
For an SVGP, the variational parameters are \(\psi = \{\mathbf{Z}, \mathbf{m}, \mathbf{S}\}\) but may also include more parameters such as kernel hyperparameters.  The approximate posterior over the function values is calculated by marginalizing over the inducing variables:
\begin{equation}
    p(\mathbf{f}_*|\mathcal{D}, \mathbf{X}_*) \approx\int_{\mathbb{R}^M} p(\mathbf{f}_*|\mathbf{u}) q(\mathbf{u}) \; d\mathbf{u}.
\end{equation}

The variational objective is derived by maximizing the Evidence Lower Bound (ELBO), which in this context becomes:
\begin{equation}
\label{eq:variationalELBOGPytorch}
L_{\text{SVGP}}(\psi| \mathcal{D}) 
= \sum^N_{i=1} \mathbb{E}_{q(f_{\text{GP}}(\mathbf{x}_i))}[\log p(y_i|f_i)] - D_{\mathbb{KL}}(q(\mathbf{u}) \parallel p(\mathbf{u}|\mathbf{Z})),
\end{equation}
where \(p(y_i|f_i)\) is the likelihood for the observations given latent function values \(f_{\text{GP}}(\mathbf{x}_i) = f_i\).

It can be shown that the complexity now becomes \(O(NM^2)\) in time and \(O(NM + M^2)\) in space \cite{pml2Book}. It is worth noting that as \(M\) increases, the approximation quality of exact inference is recovered. Too few inducing points may make the GP behave as if it was underfitting \cite{bauer2016understanding}.

\subsection{Deep Gaussian Processes and Deep Sigma Point Processes}
DGPs extend the GP framework to multiple layers, allowing for the construction of hierarchies of latent functions \cite{damianou_deep_2013, pml2Book}:
\begin{equation}
\begin{array}{l l}
\label{eq:DeepGaussianProcess}
\mathcal{DGP}(\mathbf{x}) & = \bm{f}_L \circ \cdots \circ \bm{f}_1(\mathbf{x}), \\
& \bm{f}_i(\cdot) = [f_{\text{GP}, i}^{(1)}(\cdot), \ldots, f_{\text{GP}, i}^{(H_i)}(\cdot)]^{\top}, \\
& f_{\text{GP}, i}^{(j)} \sim \mathcal{GP}(m_i(\cdot), k_i(\cdot, \cdot)).
\end{array}
\end{equation}
DGPs have a neural network-like structure with \(L\) layers, each containing \(H\) GPs. Posterior inference in GPs is no longer tractable as it requires marginalizing over a large number of RVs, corresponding to the latent function values at each layer. The stochastic variational inference method from Section \ref{sec:svgp} can be generalized to DGPs \cite{salimbeni2017doubly}, known as doubly stochastic variational inference. Each layer is agumented with inducing variables \(\mathbf{u}^{(l)}\) and corresponding variational distribution \(q(\mathbf{u}^{(l)})\). The variational ELBO for DGPs is then given by:
\begin{equation}
    L_{\text{DGP}}(\psi| \mathcal{D}) = \sum^N_{i=1} \mathbb{E}_{q(\mathbf{f}^L_i)}[\log p(\mathbf{y}_i|\mathbf{f}_i^L)] - \beta\sum^L_{l=1} D_{\mathbb{KL}}[q(\mathbf{U}^l) \| p(\mathbf{U}^l|\mathbf{Z}^{l-1})],
    \label{eq:DGPobj}
\end{equation}
where \(q(\mathbf{f}^L_i)\) denotes the approximate posterior of the final layer’s latent function for the \(i\)th data point, and \(q(\mathbf{U}^l)\) and \(p(\mathbf{U}^l|\mathbf{Z}^{l-1})\) are the variational and prior distributions over the inducing variables at layer \(l\), respectively. \(\beta > 0\) is a regularization constant. As with SVGP, the final output prediction can be integrated out analytically, however the remaining latent variables must be sampled. Resulting in `doubly` stochastic gradients from two levels of sampling: (1) data mini-batching to scale to large datasets and (2) sampling through the hidden layers (using the reparameterization trick for Gaussians) so that the latent \(\mathbf{f}_i^{(l)}\) are sampled at each layer. Note also that the expectation in \eqref{eq:DGPobj} is approximated via Monte Carlo sampling. 

DSPPs are hierarchical GP models with two key differences from DGPs \cite{jankowiak2020deep}. First, DSPPs are trained via a regularized maximum likelihood objective rather than the ELBO in \eqref{eq:DGPobj}:
\begin{equation}
    \mathcal{L}_{\text{dspp}}(\theta| \mathcal{D}) = \sum^N_{i=1} \log p_\text{dspp}(\mathbf{y}_i|\mathbf{x}_i) - \beta  \sum^L_{l=1} D_{\mathbb{KL}}[q(\mathbf{U}^l) \| p(\mathbf{U}^l|\mathbf{Z}^{l-1})],
\end{equation}
 The \(\log p_{\text{DSPP}}(\mathbf{y}_i|\mathbf{x}_i)\) directly maximizes the probability of the observed data given the model as a finite Gaussian mixture, obtained via sigma point approximations instead of Monte Carlo sampling, enabling maximum likelihood training. 
Just like DGPs the objective depends on parameters \(\theta\) containing for each GP \(\sigma_y\), \(\mathbf{m}, \mathbf{S}, \mathbf{Z}\), kernel hyperparameters and likelihood hyperparameters (e.g., observation noise), which are jointly optimized using stochastic gradient descent methods. However unlike DGPs, DSPPs also parameterize the hidden latent function values using a learnable quadrature rule. 
To make this more explicit: in DGPs, for an input \(\mathbf{x}_i\) to a layer containing \(W\) GPs, we sample:
\begin{equation}
    f_{iw} = \mu_{f_w}(\mathbf{x}_i) + \epsilon \sigma_{f_w}(\mathbf{x}_i), \quad \epsilon \sim \mathcal{N}(0,1),
\end{equation}
where \(\mu_{f_w}\) and \(\sigma_{f_w}\) are the approximate posterior mean and standard deviation predictions of the \(w\)-th GP.  
We then take \(S\) samples to obtain an unbiased Monte Carlo estimate of \(\mathbb{E}_{q(\mathbf{f}_i^L)}[\log p(\mathbf{y}_i | \mathbf{f}_i^L)]\). In DSPPs, we use the following quadrature rule:
\begin{equation}
    f_{iw}^{(j)} = \mu_{g_f}(\mathbf{x}_i) + \xi_w^{(j)} \sigma_{f_w}(\mathbf{x}_i),
\end{equation}
where \(\{\xi_w^{(j)}\}_{j=1}^{Q}\) are \(Q\) learnable quadrature points (also called \textit{sigma points}), which have associated learnable quadrature weights \(\{\omega^{(j)}\}_{j=1}^Q\). \(p_{\text{DSPP}}(\mathbf{y}_i|\mathbf{x}_i)\) is computed by evaluating the model at each of the \(Q\) quadrature sites, passing each output through the likelihood (which, for regression, is a Gaussian likelihood), and weighting them by \(\omega^{(s)}\). This produces a Gaussian mixture with \(Q\) components:
\begin{equation}
    p_{\text{dspp}}(\mathbf{y}_i|\mathbf{x}_i) = \sum^Q_{j=1} \omega^{(j)} p^{(j)}(\mathbf{y}_i|\mathbf{x}_i).
\end{equation}



   
The benefit of this is that the resulting predictive distributions are no longer degraded by posterior approximations. Only mini-batches of datapoints are sampled, so the objective is now `singly stochastic`. The intuition behind sigma points is that they represent a small, carefully selected set of points that capture the key characteristics (mean and variance) of a distribution, allowing us to approximate integrals or expectations more efficiently.

According to \citet{jankowiak2020deep}, DSPPs offer better calibrated probabilities/uncertainty quantification than DGPs, however, they only measured this using the negative log likelihood.

\subsection{Classification}

In the previous sections, we assumed that we were doing regression with a Gaussian likelihood. To adapt GPs for classification, we have to use a different likelihood to output class probabilities. In GPs, the likelihood \(p(\mathbf{y}|\mathbf{f})\) defines the relationship between the latent variable \(\mathbf{f}\) modeled by the GP and the observed data \(\mathbf{y}\). For regression and continuous outputs, the likelihood is typically a Gaussian  \(p(\mathbf{y}|\mathbf{f}) = \mathcal{N}(\mathbf{y}|\mathbf{f}, \sigma^2_y \mathbf{I})\),
which adds Gaussian noise \(\epsilon \sim \mathcal{N}(0, \sigma^2_y)\) to the predictions.

In classification, \(f_{\text{GP}}(\mathbf{x}) = f\) are the logits. Suppose we have \(D\) GP units \(\mathbf{f} = (f_1, \hdots, f_D)\) in the final layer of a hierarchical GP model. To obtain class probabilities, we can use the Softmax likelihood: 
\begin{equation}
    p(\mathbf{y}|\mathbf{f}) = \text{Softmax}(\mathbf{W} \mathbf{f}),
     \label{eq:softmaxlikelihood}
\end{equation}
where \(\mathbf{W} \in \mathbb{R}^{D \times C}\) is a learnable matrix of mixing weights applied to the latent functions \(\mathbf{f}\). This avoids needing as many GP units in the final layer as classes, which becomes computationally expensive if the number of classes is large. For our benchmarking, this was not the case, so we set \(\mathbf{W}\) to be the identity matrix.

   


\subsection{Ensembles for Uncertainty Quantification}
Ensembles \cite{NIPS2017_9ef2ed4b} provide a solid baseline for uncertainty estimation by training \(K\) independent neural networks with parameters \(\{\theta_j\}_{j=1}^K\) with randomized initializations and aggregating their predictions, which provides a Monte Carlo estimate of the Bayesian predictive posterior distribution. For regression tasks, each ensemble member is a dual-output model predicting both a mean \(\mu_i(\mathbf{x})\) and variance \(\sigma_i^2(\mathbf{x})\), trained via a Gaussian negative log-likelihood loss:  
\begin{equation}
\mathcal{L}_{\text{NN}}(\theta_j) = \sum_{i=1}^N -p_{\theta_j}(\mathbf{y}_i|\mathbf{x}_i)= \sum_{i=1}^N\frac{1}{2} \log(2\pi\sigma_j^2(\mathbf{x}_i)) + \frac{(\mathbf{y}_i - \mu_j(\mathbf{x}_i))^2}{2\sigma_j^2(\mathbf{x}_i)},
\end{equation}
which jointly optimizes accuracy and uncertainty calibration. During inference, predictions are combined into a Gaussian mixture \( \mathcal{N}(\mu_*(\mathbf{x}), \sigma_*^2(\mathbf{x})) \), where  
\begin{equation}
\mu_*(\mathbf{x}) = \frac{1}{K}\sum_{j=1}^K \mu_j(\mathbf{x}), \quad \sigma_*^2(\mathbf{x}) = \frac{1}{K}\sum_{j=1}^K \left(\sigma_j^2(\mathbf{x}) + \mu_j^2(\mathbf{x})\right) - \mu_*^2(\mathbf{x}).
\end{equation}
For classification, each member parameterizes a Gaussian distribution over logits \( \mathcal{N}(\mu_j(\mathbf{x}), \sigma_j^2(\mathbf{x})) \) and the models are trained using the cross entropy loss. Class probabilities are estimated by drawing \(K\) samples \(\hat{\mathbf{z}}_j \sim \mathcal{N}(\mu_j(\mathbf{x}), \sigma_j^2(\mathbf{x}))\), passing them through a softmax, and averaging:  
\begin{equation}
p(\mathbf{y}|\mathbf{x}) = \frac{1}{K}\sum_{j=1}^K \text{softmax}(\hat{\mathbf{z}}_j).
\end{equation}

\section{Methodology}\label{sec:methodology}

\subsection{Metrics}
\label{sec:metrics}
We used evaluation metrics, such as mean absolute error (MAE) for regression and accuracy (ACC) for classification. For uncertainty quantification, we are interested in the degree to which the model is calibrated; that is, error and misclassification should be proportional to the output uncertainty (e.g., predictive variance, max class probabilities) made by the model. So, if a classifier predicts \( p(y=c \mid x) = 0.5 \), then we expect \( c \) to be the true label 50\% of the time.

Let \(p_{\theta}(y|x)\) denote the output distribution of a predictive model, which we will assume has parameters \(\theta\).
A common class of metrics that are used to evaluate predictive models with uncertainty are proper scoring rules. A proper scoring rule is a function \(S(p_{\theta}, (y, x))\) that evaluates the quality of a predictive distribution \( p_\theta(y| x) \) against the true distribution \( p^*(y | x) \), such that the expected score is maximized when the predicted distribution equals the true distribution.
Formally:
\begin{equation}
\mathbb{E}_{(x, y) \sim p^*} [S(p_\theta, (y, x))] \leq \mathbb{E}_{(x, y) \sim p^*} [S(p^*, (y, x))],
\end{equation}
with equality if and only if \( p_\theta(y | x) = p^*(y | x) \).

The following evaluation metrics for uncertainty were used. We denote \(\downarrow\) to indicate that lower is better and vice versa for \(\uparrow\). Given a evaluation dataset \((x_i, y_i)_{i=1, \hdots, N}\) we averaged these scores over the samples:

\textbf{Negative Log-Likelihood (NLL)} \(\downarrow\): A proper scoring rule, although it can overemphasize tail probabilities \cite{watkins1992q}. This is the only evaluation metric used for uncertainty in \citet{jankowiak2020deep}. For classification, we used a categorical likelihood:
 \begin{equation} S_{\text{NLL}}(p_\theta, (y, x)) = -\log p_\theta(y | x), \end{equation}
and for regression the Gaussian likelihood:
 \begin{equation} S_{\text{NLL}}(p_\theta, (y, x)) = \frac{1}{2} \log(2\pi \sigma^2(x)) + \frac{(y - \mu(x))^2}{2 \sigma^2(x)},\end{equation} where \(\mu(x)\) and \(\sigma^2(x)\) are the model-predicted mean and variance for input \(x\).


\textbf{Expected Calibration Error (ECE)} \(\downarrow\): Not a proper scoring method, but useful for assessing calibration. 
It is related to reliability plots. In case of classification we look at \(B\) bins each with indices \(\mathcal{B}_b = \{i \in [1, N] : p_{\theta}(y_i|x_i) \in (\frac{b-1}{B}, \frac{b}{B}]\}\) such that each bin has predictions whose confidence fall within a confidence interval. We can compute the accuracy within bin \(b\) as 
\begin{equation}
    \text{acc}(\mathcal{B}_b) = \frac{1}{|\mathcal{B}_b|} \sum_{n \in \mathcal{B}_b} \mathbb{I}(\hat{y}_n=y_n),
\end{equation}
where \(\hat{y}_n\) is the max probability class and \(y_n\) the true class.
We then define \(\text{conf}(\mathcal{B}_b)\) as the average confidence (predictive variance or max probability) within a bin and plot the accuracy against the confidence, obtaining the reliability plot. The ECE is computed as:
\begin{equation}
    \text{ECE} = \sum^B_{b=1} \frac{|\mathcal{B}_b|}{B} |\text{acc}(\mathcal{B}_b) - \text{conf}(\mathcal{B}_b)|.
\end{equation}
 For regression, we can consider confidence interval accuracy instead, and a set of equally spaced confidences \(\alpha \in S_{\alpha}\) dividing \([0,1]\) into \(B\) equally spaced values. For each prediction interval at confidence level \(\alpha\), the confidence interval accuracy is the fraction of true target values \(y\) falling within the interval:
 \begin{equation}
     \text{acc}(\alpha) = \frac{1}{N} \sum^N_{i=1} \mathbb{I}(y_i \in [l_i, u_i]),
 \end{equation}
where \(l = \mu - |z_{\frac{\eta}{2}}|\sigma, u = \mu + |z_{\frac{\eta}{2}}|\sigma\) with \(\eta=1+\alpha\) and \(z_{\frac{\eta}{2}}\) being the z score corresponding to the \(\eta/2\) quantile.
 
Calibration error is then calculated analogously to the classification case for expected calibration error:
 \begin{equation}
     \text{CE}_{\text{reg}} = \frac{1}{B} \sum^B_{i=1} |\alpha_k - \text{acc}(\alpha_k)|.
 \end{equation}


\subsection{Datasets}


We evaluate our models on two distinct tasks using publicly available datasets: regression on protein structure data and classification on epileptic seizure recognition data.

\textbf{Physicochemical Properties of Protein Tertiary Structure (CASP):}
For the regression task, we used the dataset detailing the physicochemical properties of protein tertiary structure, sourced from CASP 5-9 and originally compiled by Prashant Rana \cite{protein}. This dataset consists of 45,730 protein decoys. The objective is regression analysis to predict a continuous variable ranging from 0 to 21 Angstroms. The input features comprise 9 numerical attributes (\texttt{F1} through \texttt{F9}) that describe various physicochemical properties of a protein. All features are real-valued. For our experiments, we employed an 80:20 split for training and testing, respectively. The features were standardized (zero mean, unit variance) before training.

\textbf{Epileptic Seizure Recognition (ESR):}
For the classification task, we utilized the Epileptic Seizure Recognition dataset \cite{esr}. This dataset contains 11,500 instances, each representing a one-second segment of EEG recordings. It includes 178 numerical features (\texttt{X1} through \texttt{X178}) corresponding to EEG signal values at different time points. The original dataset is formulated as a 5-class classification problem, where class 1 represents seizure activity, and classes 2-5 represent various non-seizure states (e.g., tumor area, healthy area with eyes open/closed). For our study, we converted this into a binary classification task: predicting the presence (class 1) versus the absence (classes 2-5 merged into class 0) of a seizure. The features were standardized (zero mean, unit variance) before training. Similar to the regression task, we used an 80:20 train/test split for evaluation.

\subsection{Hyperparameter Tuning}
We performed hyperparameter optimization using Bayesian optimization (BayesOpt). See Table \ref{tab:hyperparams} for the set of hyperparameters we used and the range of their values. Some hyperparameters we fixed, while others were tunable by BayesOpt. To prevent data leakage, we further divide the training set into an 80:20 ratio for validation during hyper-parameter tuning. As a reminder, the BayesOpt algorithm proceeds as follows: We start with an initial dataset \(\mathcal{D}_0= (\mathbf{x}_i, y_i)_{i=1,...,n_0}\) where the target outputs \(y_i = f(\mathbf{x}_i) + \epsilon_i\) are assumed to be noisy outputs of the function \(f\) we want to optimize and \(\mathbf{x}_i\) are sobol (quasi-random, low-discrepancy) points in the input space. Afterwords, at each iteration \(n\), a dataset \(\mathcal{D}_n\) is maintained. A GP\footnote{This is a separate GP from the (deep) GP that is fit on the train set in case we are optimizing the hyperparameters of a (deep) GP. Furthermore, this GP uses exact inference instead of a variational approximation.} can then be used to estimate \(p(f|\mathcal{D})\), a distribution over \(f\). An acquisition function \(\alpha(\mathbf{x}; \mathcal{D}_n)\) is then used to select a new candidate \(\mathbf{x}\) based on its expected utility. Once \(y_{n+1} = f(\mathbf{x}_{n+1}) + \epsilon_{n+1}\) has been observed, the GP is updated by computing \(p(f|\mathcal{D}_{n+1})\). 

For hyperparameter tuning we used the expected improvement acquisition function and ran the optimization loop for 20 trials, 5 of which were used for random initialization. We optimized the negative log likelihood \(S_{\text{NLL}}\)
as it is a proper scoring rule. 

For the remaining experimental setup sections, we used the optimal hyperparameters that were obtained for a particular dataset and model after BayesOpt.

\begin{table}[ht]
\centering
\caption{Hyperparameter Setup for Model Training: Kernel (RBF) and likelihood (Gaussian or Softmax) were shared for GP-based models. And for NN we used ReLU activation functions. Learning rates are optimized on a log scale.  Notation for layers: \([n_0, \hdots , n_L]\) where \(L\) is the number of hidden layers and \(n_l\) is the number of units in layer \(l\). \([\;]\) is used to indicate no hidden layers.}
\label{tab:hyperparams}
\begin{tabular}{@{}lll@{}}
\toprule
\textbf{Hyperparameter} & \textbf{Value or Range} & \textbf{Applicable Models} \\ 
\midrule
Monte Carlo samples  & Fixed: 10 & DGP \\  
Quadrature sites & Fixed: 8 & DSPP \\  
Scaling of KL divergence \((\beta)\) & Fixed: 1 & DGP/DSPP \\ 
Epochs & Fixed: Until convergence   & All \\  
& (CASP: 20, ESR: 30) & \\
Learning rate & Varied: $10^{-3}$–$10^{-1}$ (log) & All \\  
Model architecture (layers) & Varied: \([\;], [1], [1, 1], [3], [3, 3]\) (Protein) & DGP, DSPP \\  
& Varied:  \([\;], [1], [1, 1], [5], [5, 5]\) (ESR) & DGP, DSPP \\
Number of inducing points ($M$) & Varied: 50–200 & DGP, DSPP \\ 
NN architecture (layers/units) & Varied: \([2n, n], n \in [8,16,32,64]\) & Deep Ensemble \\  
\# Models & Varied: 2-10 & Deep Ensemble \\
Optimizer & Adam & All \\
\bottomrule
\end{tabular}
\vspace{0.2cm}
\small
\\
\end{table}


\subsection{Experiment Setup}
Here, we describe the experiment conducted to obtain the final evaluation results. The same procedure was applied to both the regression (Protein) and classification (ESR) datasets.

We trained three models—DGPs, DSPPs, and an ensemble of MLPs—on the training split and evaluated them on the test split using the metrics covered in Section \ref{sec:metrics}, namely NLL, ECE, accuracy and MAE.
This includes producing a calibration plot showing the confidence of the models against their accuracy. We also kept track of training and validation loss curves to assess model fit.


\subsection{Distribution Shift Experiment Setup}
\label{sec:preturbation}
To evaluate the robustness and calibration of our models under inputs that deviate from the training distribution, we designed a synthetic feature-level shift framework that can be applied uniformly across regression and classification tasks. This enables a controlled, systematic stress test of model performance and uncertainty quantification beyond in-distribution settings.  
We define five classes of perturbations, each parameterized by a severity level \(s \in [0, 1]\). Let \(\mathbf{X} = (\mathbf{x}_1, \hdots, \mathbf{x}_N)\) denote the data matrix of feature vectors:

\textbf{Gaussian Noise}
\begin{equation} \mathbf{X}' = \mathbf{X} + \varepsilon, \quad \varepsilon \sim \mathcal{N}(0, (\sigma s)^2). \end{equation}
We add Gaussian Noise to the features: the corrupted features \(\mathbf{X}'\) are obtained by adding noise \(\varepsilon\) to the original features \(\mathbf{X}\), where the noise is drawn from a zero-mean normal distribution with standard deviation \(\sigma s\).

\textbf{Feature Masking}
\begin{equation}
\mathbf{X}'_{i,j} = \begin{cases} 
0 & \text{with probability } s, \\
\mathbf{X}_{i,j} & \text{otherwise}.
\end{cases}
\end{equation}
We simulate missing or occluded features by randomly setting each feature value to zero with probability \(s\), leaving it unchanged otherwise.

\textbf{Feature Scaling}
\begin{equation}
\mathbf{X}' = (1 + s) \cdot \mathbf{X}.
\end{equation}
We apply a global scaling factor \(1 + s\) to all features, modeling scenarios such as unnormalized sensor inputs or calibration mismatches. 

\textbf{Feature Permutation}
\begin{equation}
\mathbf{X}'_{i,j} = \begin{cases}
    \mathbf{X}_{\pi(i), j} & \text{with probability } s, \\
    \mathbf{X}_{i, j} & \text{otherwise}.
\end{cases}
\end{equation}
Each feature value \(\mathbf{X}_{i,j}\) is randomly replaced with another value from the same column (row \(\pi(i)\)) with probability \(s\), disrupting inter-feature dependencies while preserving marginal distributions.

\textbf{Outlier Injection}
\begin{equation}
\mathbf{X}'_{i,j} = \mathbf{X}_{i, j} + \delta, \quad \delta \in \{\pm 3 \cdot \sigma_j \} \text{ with probability } s.
\end{equation}
Outliers are injected by perturbing a fraction \(s\) of the features with large positive or negative deviations proportional to the pre-feature standard deviation \(\sigma_j\).

To assess robustness under distribution shift, we performed \(N = 5\) independent training runs per model, each initialized with different random seeds. In each run, the model is re-initialized, trained from scratch, and subsequently evaluated on test sets perturbed by all corruption types at increasing severity level \(s \in \{0.0, 0.1, 0.2, 0.4, 0.6, 0.8\}\). 

At each severity level, we computed ECE, accuracy and MAE for classification and regression, respectively. Results were aggregated across all runs and five perturbation types, yielding 25 metric values per severity. These are visualized using boxplots to reflect variation due to both training stochasticity and corruption type. As models are evaluated without retraining on the shifted test sets, the results quantify robustness to distributional changes that preserve overall structure while altering feature-level statistics.

Our shift analysis framework builds on work in robustness and uncertainty evaluation under corruptions, covariate shift, and non-adversarial perturbations \cite{hendrycks2019benchmarking, geirhos2018generalisation, ilyas2019adversarial, hendrycks2019augmix, ovadia2019trustmodelsuncertaintyevaluating}, and complements regularization-based approaches to improving model robustness \cite{srivastava2014dropout}.

\section{Results}\label{sec:Results}

This section presents the empirical evaluation of our models on the selected datasets. The experiments utilized the optimal hyperparameters identified via Bayesian optimization, as detailed in Appendix A (Table \ref{tab:hyper-params}). We evaluate the models on regression (CASP dataset) and classification (ESR dataset) tasks, focusing on predictive performance, uncertainty calibration, and robustness under distribution shift.

\begin{figure}[h!]
    \centering
    \includegraphics[width=0.48\linewidth]{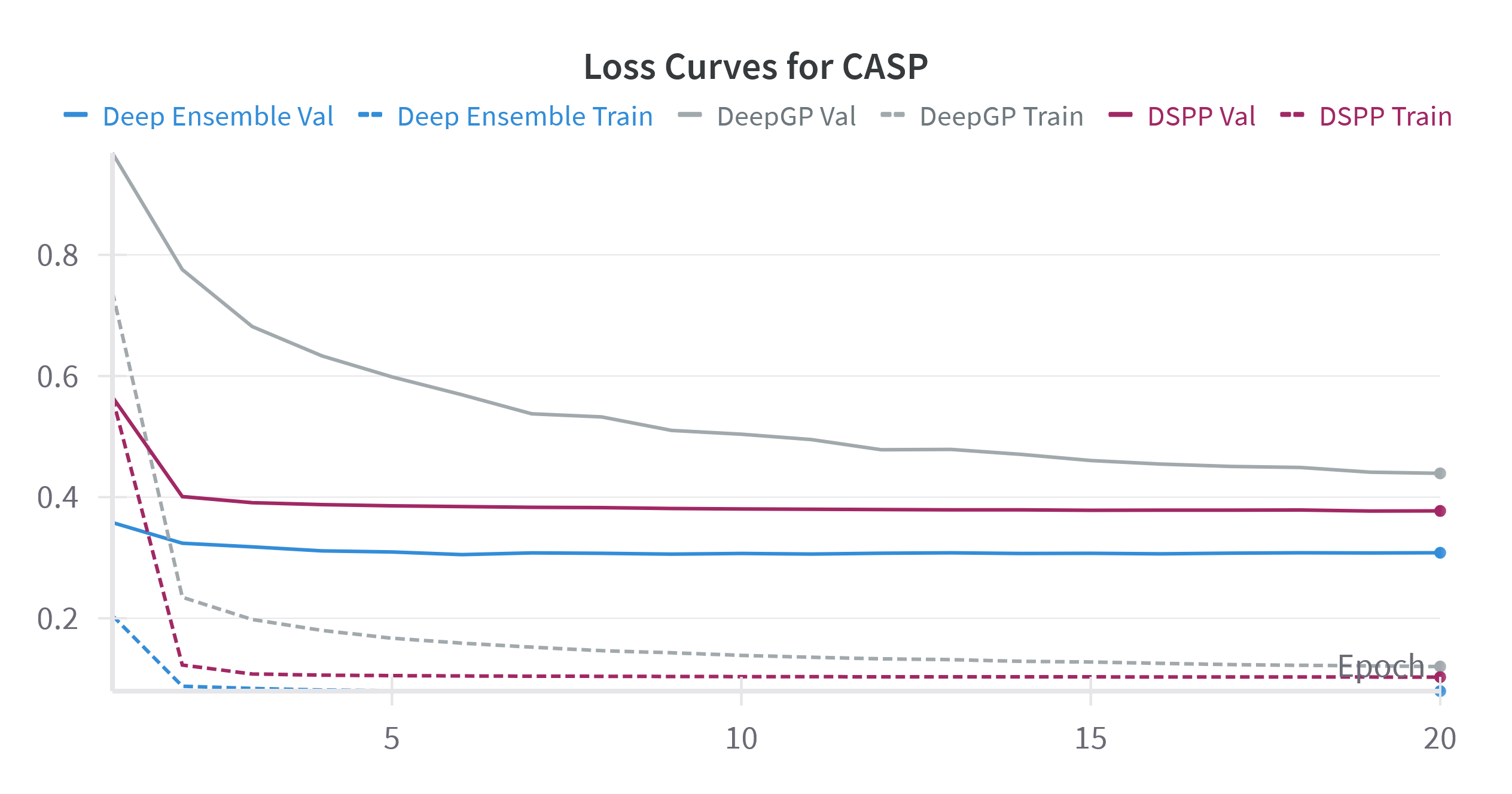}
    \includegraphics[width=0.48\linewidth]{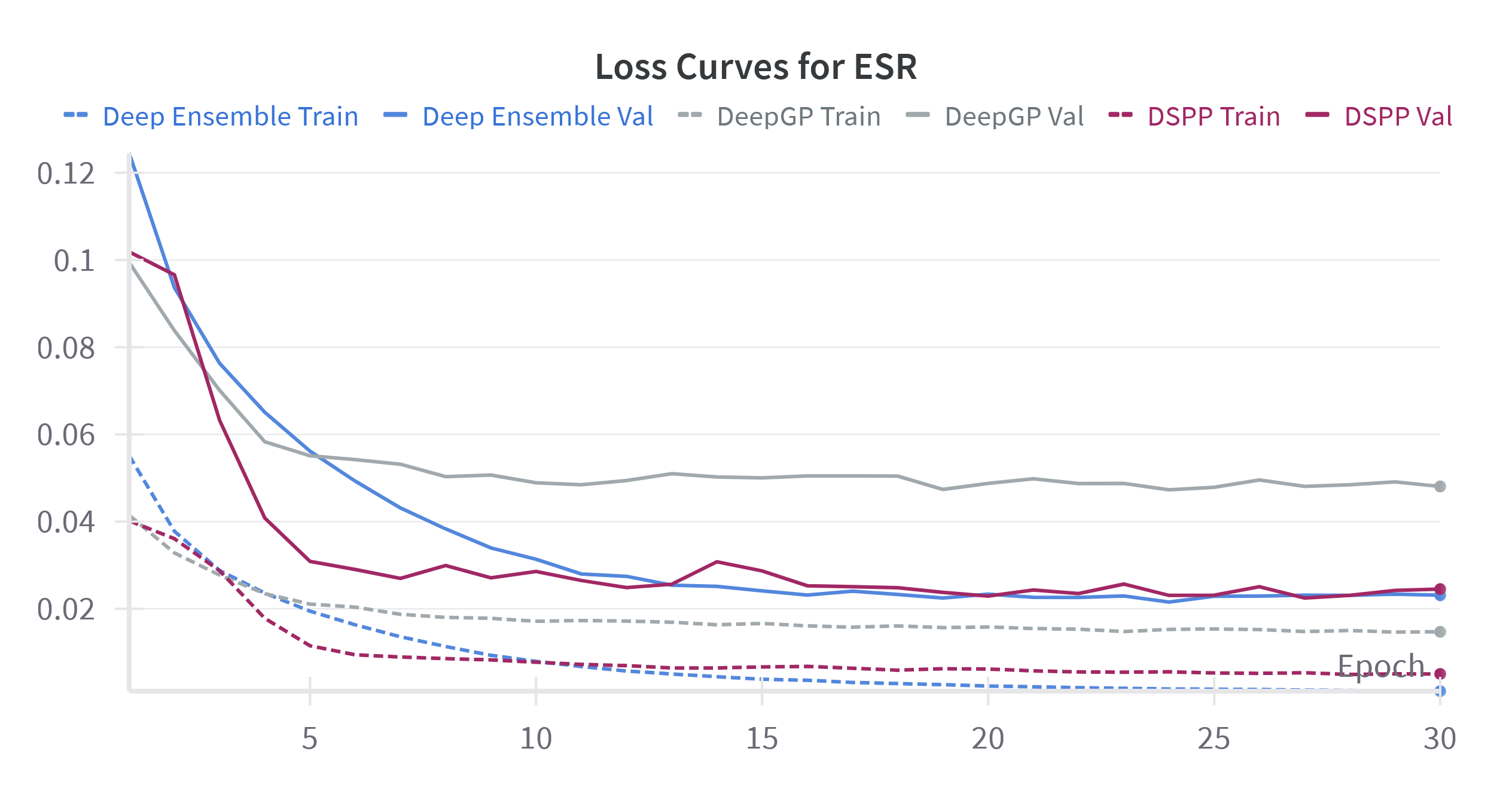}
    \caption{Training and validation loss curves of our optimized models on CASP and ESR}
    \label{fig:loss-curves}
\end{figure}

An examination of Figure \ref{fig:loss-curves} reveals that all models demonstrate expected loss curves, without significant overfitting to the training set. However, the loss gap between train and validation sets can still be remedied using regularization techniques like dropout and weight penalties. Additionally, in the case of DSPPs, the $\beta$ hyper-parameter can be lowered to reduce the constraint that the approximate posterior should not deviate too much from the prior.

Table \ref{tab:results} summarizes the primary performance metrics on the respective test sets. For the CASP regression task, DSPP demonstrated superior performance in terms of uncertainty quantification, achieving the lowest NLL of 2.985 and the best ECE of 0.026. While DSPP excelled in calibration, the DGP model obtained the lowest MAE of 3.425, indicating the most accurate point predictions on average for this task. The Deep Ensemble baseline was outperformed by both GP variants in NLL and MAE, although its ECE (0.112) was slightly better than that of the DGP (0.132).

The results on the ESR classification task reveal a different trend between models. The Deep Ensemble baseline achieved the highest accuracy at 0.976 and the lowest NLL (0.073), suggesting strong predictive performance. However, DSPP provided the best calibration with the lowest ECE (0.035) while maintaining a high accuracy of 0.969. The DGP model lagged behind the other two methods in both accuracy (0.912) and NLL (0.258) on this classification benchmark, although its calibration (ECE 0.054) was not substantially higher than its competitors.

\begin{figure}[h!]
    \centering
        \includegraphics[width=0.48\textwidth]{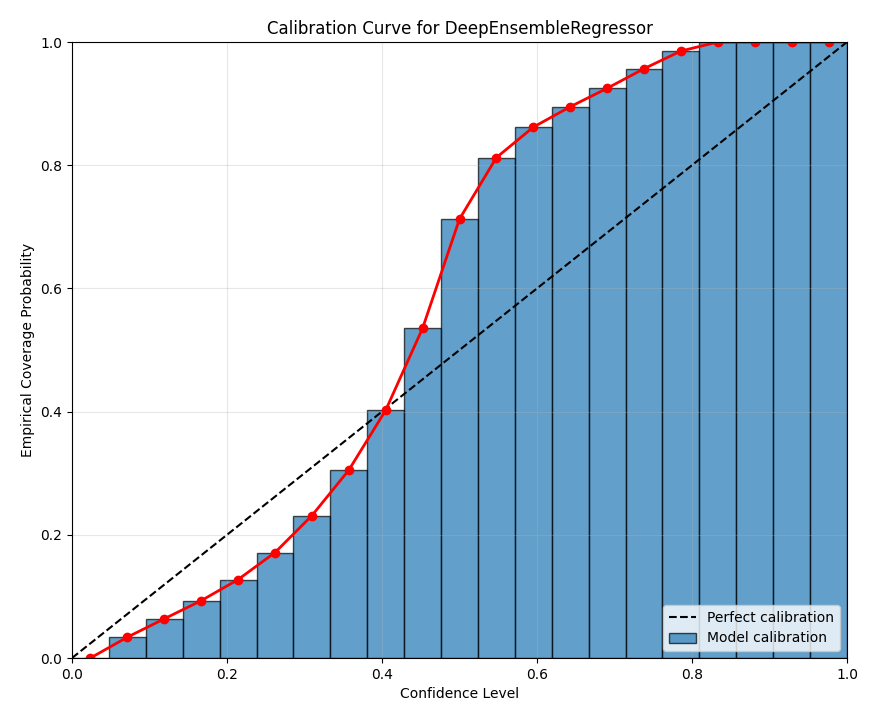}
        \includegraphics[width=0.48\textwidth]{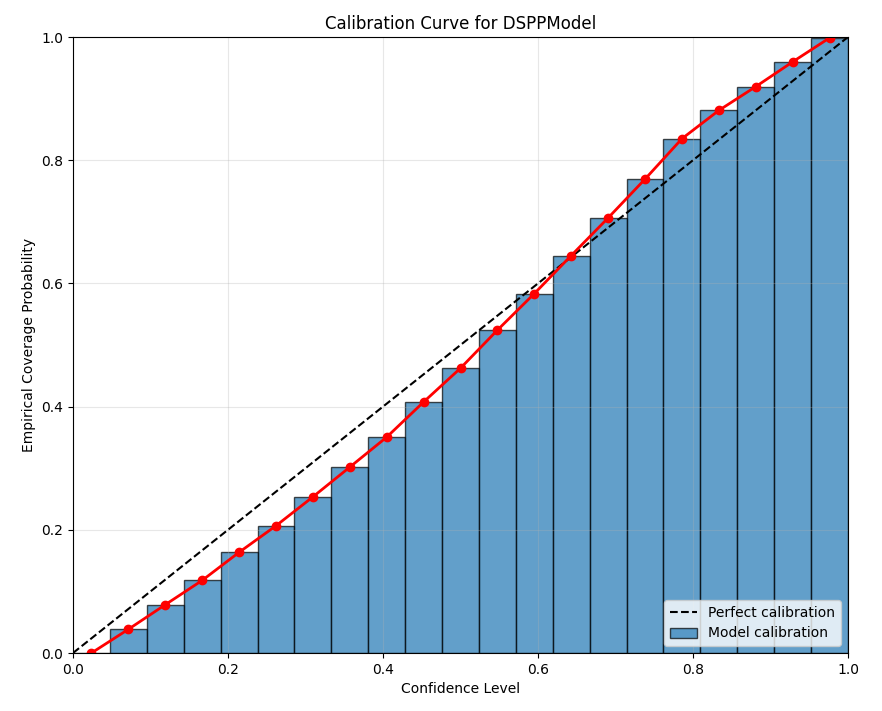}
        \includegraphics[width=0.48\textwidth]{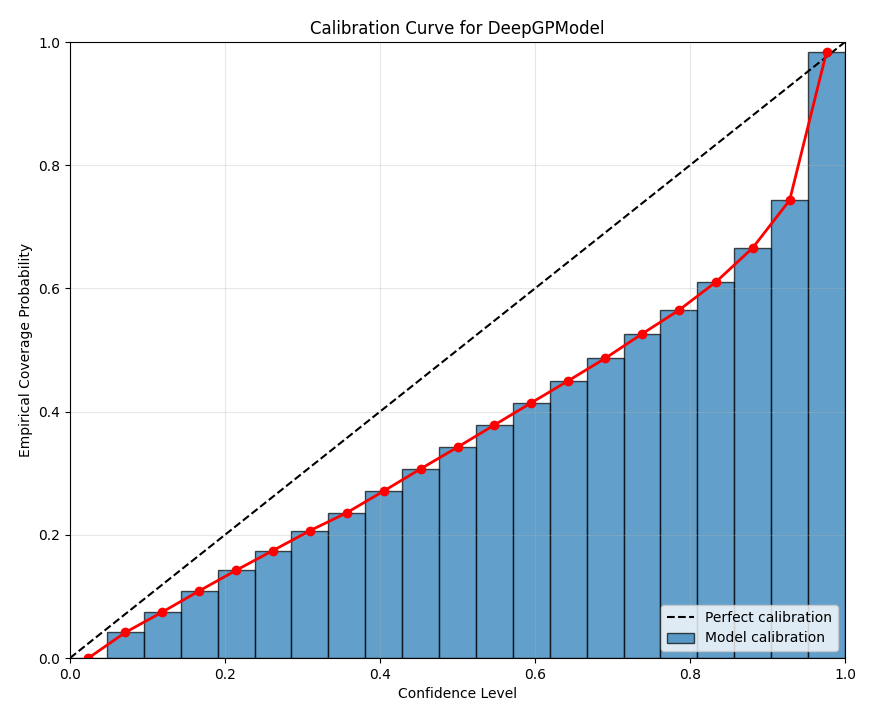}
    \caption{Calibration curves of our models for the Protein regression dataset.}
    \label{fig:mainfig}
\end{figure}

Visual inspection of the calibration curves provides further insight into model confidence. Figure \ref{fig:mainfig} displays the reliability plots for the CASP regression task. The DSPP model's curve closely tracks the ideal diagonal line, visually confirming its excellent calibration reported in Table \ref{tab:results} and indicating neither significant over nor underconfidence. The DGP model shows a curve consistently lying below the diagonal across nearly all confidence levels, indicating a clear tendency towards overconfidence; the model predicts higher certainty than its accuracy justifies. The Deep Ensemble model displays a more mixed pattern: it appears slightly overconfident at low confidence levels but becomes increasingly underconfident at mid-to-high confidence levels.

\begin{figure}[h!]
    \centering
        \includegraphics[width=0.48\textwidth]{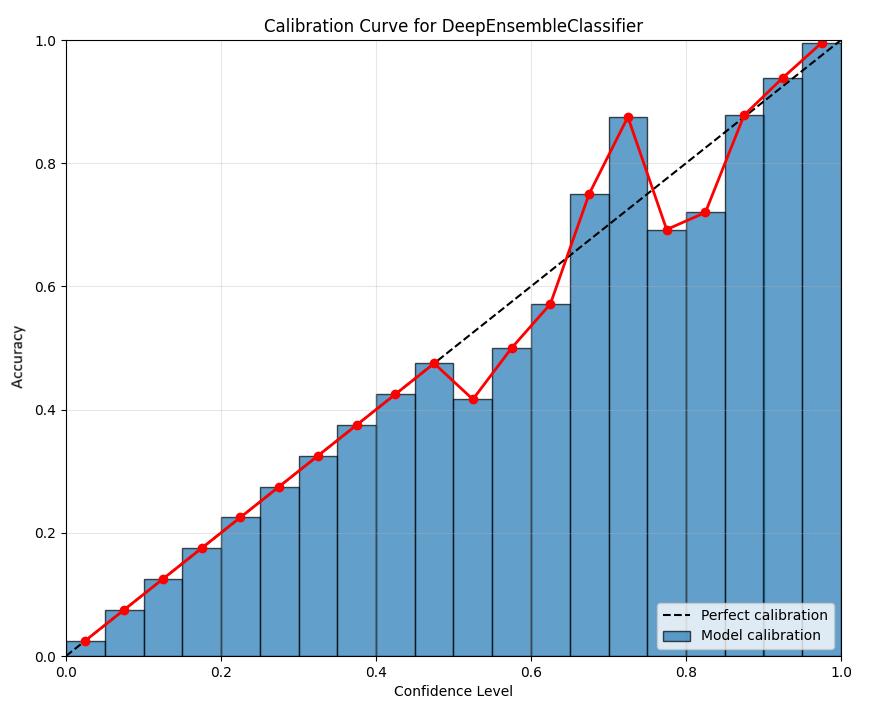}
        \includegraphics[width=0.48\textwidth]{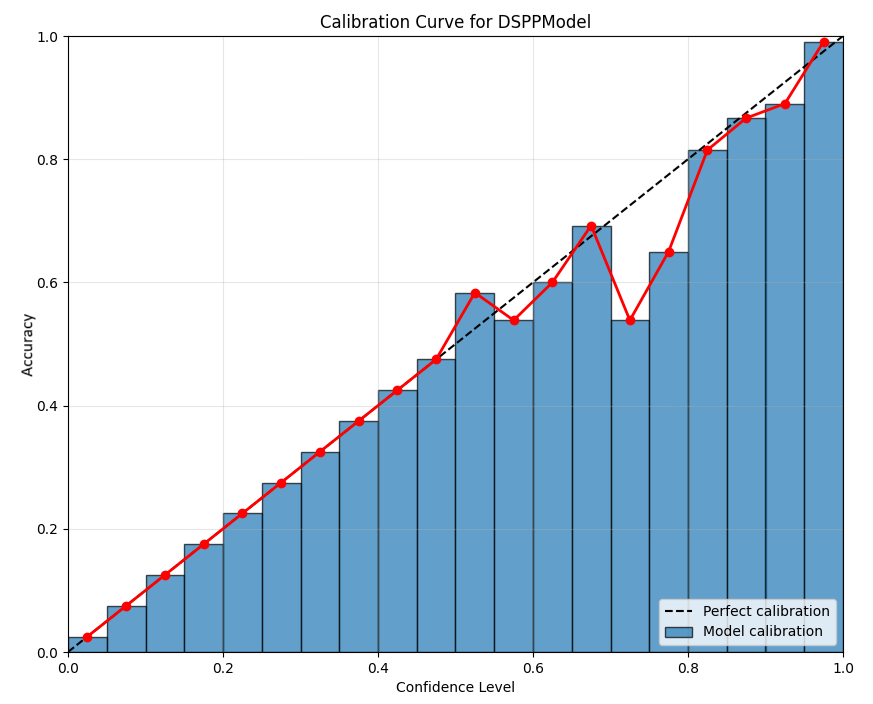}
        \includegraphics[width=0.48\textwidth]{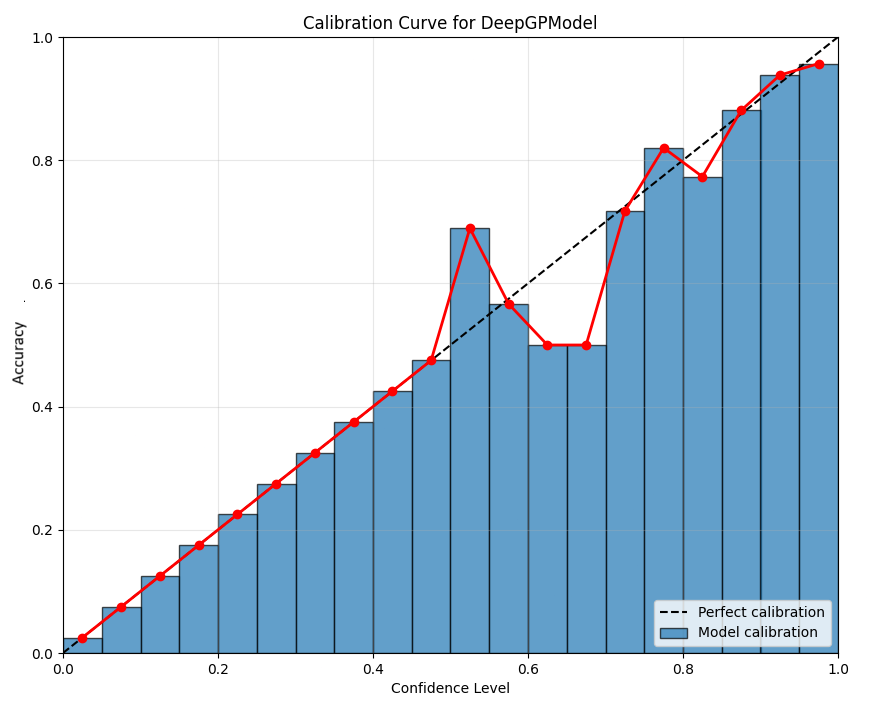}
    \caption{Calibration curves of our models for the Epileptic Sezure Recognition dataset.}
    \label{fig:mainfig2}
\end{figure}

For the ESR classification task (Figure \ref{fig:mainfig2}), all three models demonstrate good calibration. The calibration curves for the Deep Ensemble, DSPP, and DGP all closely follow the ideal diagonal line, indicating that the models are generally well-calibrated. The visual differences between the plots are minor, aligning with the strong quantitative ECE performance observed for all models on this dataset (Table \ref{tab:results}).

To assess performance under data distribution shifts, we evaluated the models on perturbed test sets using the methodology described in Section \ref{sec:preturbation}. The results are summarized through grouped boxplots, where each severity level aggregates performance across all types of corruptions. 

\begin{figure}[h!]
    \centering
    \includegraphics[width=0.48\linewidth]{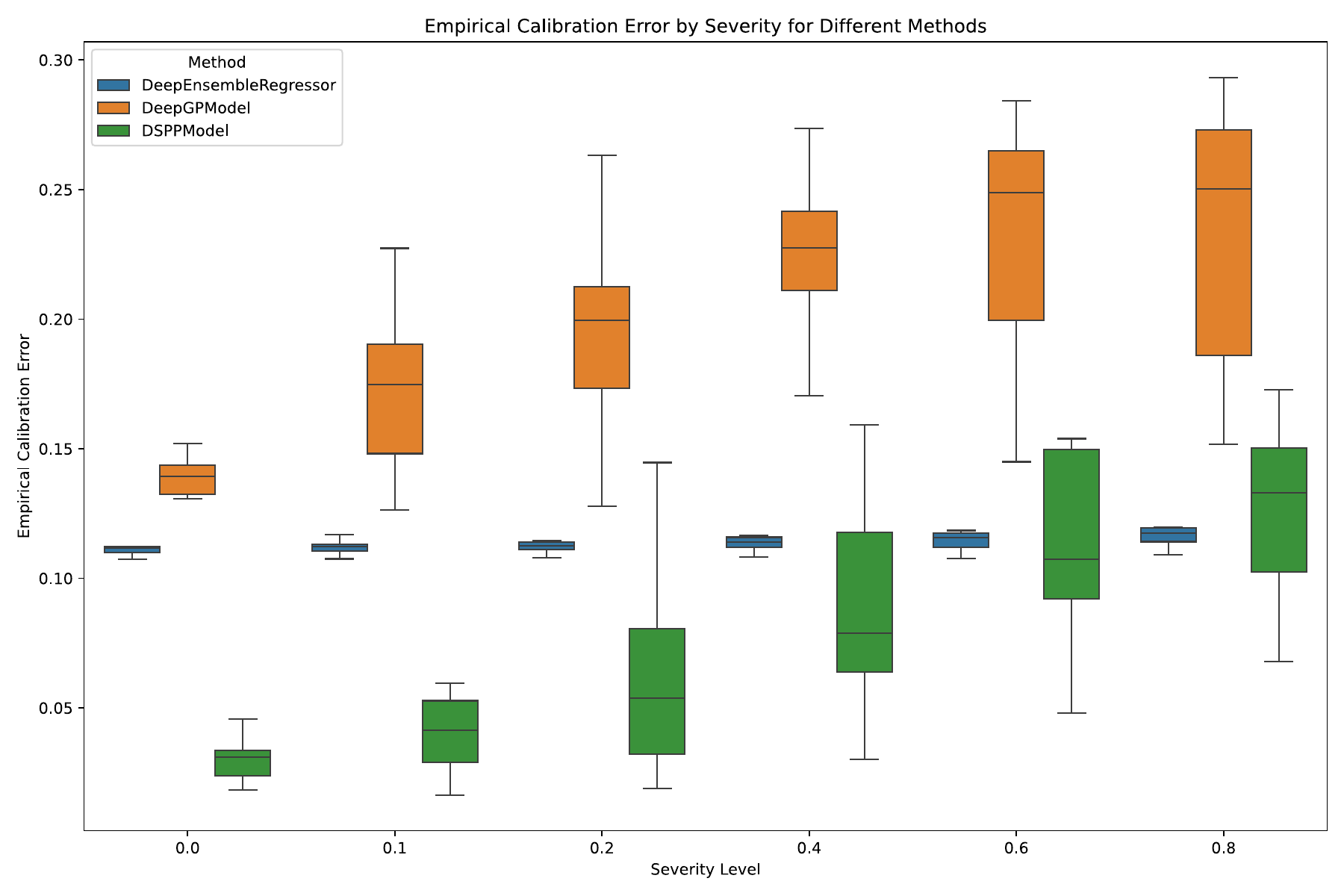}
    \includegraphics[width=0.48\linewidth]{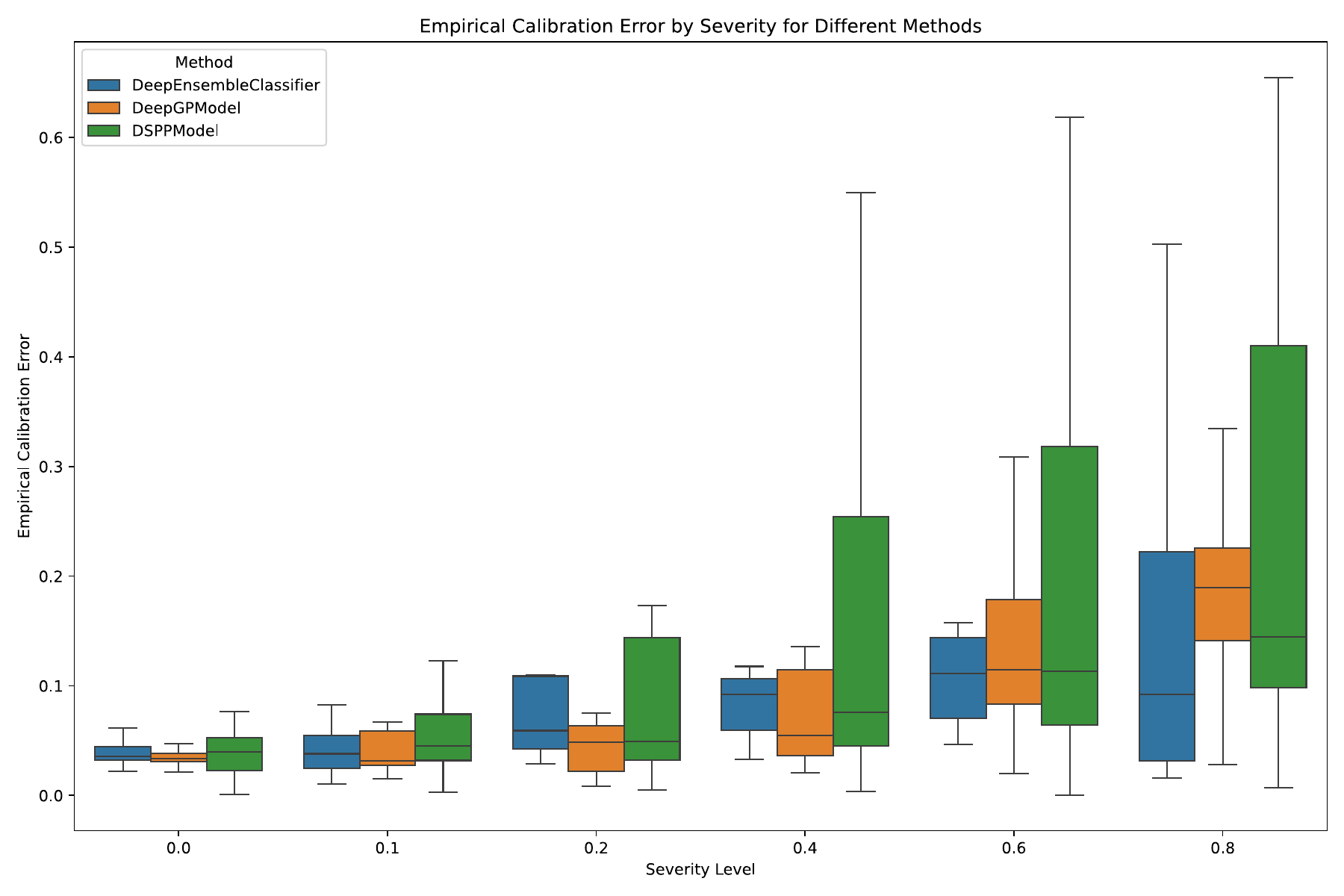}
    \caption{Box plots showing ECE under distributional shift for the CASP regression (left) and ESR classification (right) tasks across three methods: Deep Ensemble (blue), DGP (orange), and DSPP (green).}
    \label{fig:boxplotECE}
\end{figure}

Figure \ref{fig:boxplotECE} shows the ECE under increasing shift severity for the CASP regression task. While both DGP and DSPP exhibit rising ECE with severity, the Deep Ensemble remains stable around 0.11. DGP shows the highest ECE and is most affected by shift, whereas DSPP maintains the lowest ECE across most severity levels and increases more gradually. Overall, DSPP demonstrates the strongest calibration performance under distributional shift. 

Turning to the ESR classification task calibration trends on ESR are illustrated in Figure \ref{fig:boxplotECE}. All three models maintain low ECE values across the full range of shift severities, indicating generally strong calibration performance. While DSPP continues to exhibit competitive calibration with the lowest ECE median at most levels, its variability increases under stronger shifts. The Deep Ensemble also maintains low calibration error but shows a slightly broader distribution at the highest severity. Notably, DGP displays stable calibration with relatively narrow interquartile ranges, though its median ECE is slightly higher than the others in the later stages. 

Finally, ablation studies were conducted to understand the impact of key hyperparameters (Appendix \ref{sec:ablation}). The results, shown in Figures \ref{fig:nllbruh} and \ref{fig:nll}, indicate that model performance (measured by NLL) is sensitive to both the number of inducing points \(M\) and the model depth (number of layers). Generally, increasing \(M\) improved NLL up to a certain point for both DGP and DSPP (Figure \ref{fig:nllbruh}), highlighting the trade-off between approximation quality and computational cost inherent in sparse GP methods. The effect of depth was dataset-dependent (Figure \ref{fig:nll}); for CASP, NLL tended to increase with more layers for DSPP, while for ESR, deeper models consistently performed better. 


\begin{table}[htbp]
  \centering
  \caption{Performance metrics for our optimized models on their corresponding datasets.}
  \label{tab:results}
  \begin{tabular}{@{} l l c c c c @{}}
    \toprule
    \textbf{Model} & \textbf{Dataset} 
      & \textbf{NLL $\downarrow$} 
      & \textbf{ECE $\downarrow$} 
      & \textbf{MAE $\downarrow$} 
      & \textbf{ACC $\uparrow$} \\
    \midrule
    DGP & CASP & 3.412 & 0.132 & \textbf{3.425} & --- \\
    Deep Ensemble & CASP & 3.387 & 0.112 & 5.525 & --- \\
    DSPP & CASP & \textbf{2.985} & \textbf{0.026} & 4.016 & --- \\
    \midrule
    DGP & ESR & 0.258 & 0.054 & --- & 0.912 \\
    Deep Ensemble & ESR & \textbf{0.073} & 0.040 & --- & \textbf{0.976} \\
    DSPP & ESR & 0.091 & \textbf{0.035} & --- & 0.969 \\
    \bottomrule
  \end{tabular}
\end{table}

\section{Discussion}\label{sec:discussion}


Our results highlight the potential of deep probabilistic models in trustworthy machine learning. In particular, we observe that GP-based models (DGPs and DSPPs) provide well-calibrated uncertainty estimates as well as good predictive performance, but this advantage may vary depending on the dataset and task type. We confirm empirical results from \citet{jankowiak2020deep} that the use of deterministic approximation (via sigma points) to propagate uncertainty through nonlinear transformations provides better calibrated uncertainty estimates compared to the standard sampling-based approach in DGPs (which also makes the model more computationally efficient). Furthermore, our experiments demonstrate the effectiveness of this approach for non-Gaussian likelihoods (specifically, the Softmax likelihood in the ESR classification task), an aspect not previously explored empirically for DSPPs.

A crucial finding emerged from the distribution shift experiments. Deep Ensembles displayed notable resilience, maintaining relatively stable performance and calibration across increasing perturbation severities (Figures \ref{fig:boxplotECE}, \ref{fig:boxplotMAE&ACC}). In contrast, the GP-based models showed greater sensitivity. DGPs were particularly affected on the regression task, with significant degradation in both prediction accuracy (MAE) and calibration (ECE). DSPPs presented a more nuanced picture under shift: on the CASP regression task, they maintained the best calibration (lowest median ECE) despite the increasing shift, outperforming DGPs significantly in this regard. However, on the ESR classification task, while competitive at low severities, DSPP accuracy degraded more sharply than ensembles, and its ECE showed increasing variability under stronger shifts, indicating less reliable calibration compared to ensembles in that scenario. This suggests that while the sigma point method aids calibration robustness in some cases, the overall model structure's resilience to feature perturbations might be lower than that of ensembles, depending on the task and dataset. This underscores that good in-distribution calibration does not guarantee robustness, necessitating evaluation under shift.

Our study, however, has limitations. The focus on moderately complex, tabular datasets restricts the generalizability of these findings to high-dimensional data common in fields like computer vision or NLP, where architectural differences might significantly influence robustness. Furthermore, while DSPPs avoid sampling during inference, potentially offering speed advantages over DGPs, we did not perform rigorous wall-clock time comparisons. Meaningful assessment of computational efficiency requires strictly controlled and standardized environments to ensure fair comparisons free from confounding system variables.

\subsection{Future Work}

In this work we focused on standard likelihoods for regression and classification with deep GPs. Future work could expand upon this by including a larger class of likelihoods.  For regression tasks, Laplace or Student-\(t\) likelihoods can provide greater resilience to outliers due to their heavier tails. For classification, Bernoulli and categorical likelihoods are standard choices, but Ordinal or Dirichlet-based likelihoods may be better suited for tasks involving ranked or multi-class uncertainty.
A Poisson likelihood could be used for Poisson regression to model count data where outcomes represent the number of events occurring in a fixed interval. In this scenario the underlying log rate function is modeled by a GP. More generally, a Poisson process where the log rate function is itself a stochastic process is known as a `Cox process` or `doubly stochastic Poisson process` \cite{10.1111/j.2517-6161.1955.tb00188.x, pml2Book}.

It is also worth re-exploring the idea of deep kernel learning \cite{pmlr-v51-wilson16} but for DGPs/DSPPs. Deep kernel learning involves combining neural networks with GPs by using a neural network as a feature extractor before applying the GP. In other words, the GP operates on the latent space produced by a neural network as opposed to the raw input space. Previous work in this area primarily focused on deep kernel learning with a single GP. While DGPs/DSPPs can, in theory, learn hierarchical feature representations, their MLP-like structure makes them less suitable for certain data types. For example, it is difficult to scale these models to image data due to the large number of features corresponding to the pixels in the image. A potential solution is to use a convolutional neural network or vision transformer followed by the deep GP, which can then be trained in an end-to-end fashion.

This work focused on standard on the standard regression and classification. However a promising area of research is the use 
of GPs for uncertainty quantification in Reinforcement Learning (RL) \cite{engel2005reinforcement, chowdhary_off-policy_2014, pmlr-v32-grande14, kameda2023reinforcement, pmlr-v265-lende25a}. The use of uncertainty quantification in RL has several use cases \cite{lockwood2022review}. RL algorithms that quantify (epistemic) uncertainty could address sample complexity through more intelligent exploration of the environment by letting the agent focus on high-uncertainty regions likely to yield informative transitions. The field of safe RL \cite{berkenkamp2019safe} uses uncertainty quantification to adhere to safety constraints during the learning process, which is important in robotics applications, where overly aggressive exploration could potentially damage the physical system.
Recently, 
\citet{pmlr-v265-lende25a} explored using DGPs for estimating the action-value function, quantifying the uncertainty of actions in a state in addition to its expected value.
However, this framework has not been extended to DSPPs or to estimating the policy directly.

\section{Conclusion}\label{sec:conclusion}

Our experiments demonstrate that deep Gaussian process models can deliver competitive predictive performance while providing robust and well-calibrated uncertainty estimates. The results suggest that DSPPs often outperform DGPs and ensembles in terms of calibration, showing strong reliability across varied tasks and moderate distributional shifts. These findings highlight the promise of DSPPs for trustworthy AI solutions, where model confidence is as important as predictive accuracy. Future research can explore wider applications, include more complex likelihoods, and further examine how these methods adapt to larger and more diverse datasets.

\bibliographystyle{plainnat} 
\bibliography{main}

\clearpage
\appendix

\section{Optimized Model Hyper-parameters}

We report the results of our hyper-parameter tuning for each combination of model and dataset in Table \ref{tab:hyper-params}. Besides the variables mentioned in the table, we fix the number of quadrature points in DSPP to 8 and the number of Monte Carlo samples in DGP to 10.

\begin{table}[h]
    \centering
    \caption{Optimized hyper-parameters after Bayesian search.}
    \begin{tabular}{@{} l l c c c c c @{}}
    \toprule
    \textbf{Model} & \textbf{Dataset} & \textbf{LR} & \textbf{\# Epochs} & \textbf{Arch} & \textbf{\# Inducing Points} & \textbf{\# Models} \\
    \midrule
    DGP & CASP & 0.1 & 20 & [3] & 159 & --- \\
    Deep Ensemble & CASP & 0.025 & 20 & [128, 64] & --- & 9 \\
    DSPP & CASP & 0.055 & 20 & [\;] & 50 & --- \\
    DGP & ESR & 0.1 & 30 & [5, 2] & 200 & --- \\
    Deep Ensemble & ESR & 0.001 & 30 & [128, 64] & --- & 10 \\
    DSPP & ESR & 0.068 & 30 & [5, 5, 2] & 50 & --- \\
    \bottomrule
    \end{tabular}
    \label{tab:hyper-params}
\end{table}

\section{Variational Inference}
\label{sec:vi}
To make training scalable, GP based methods make use of variational inference (VI). We briefly cover here how VI works in general.

VI is a method for approximating complex probability distributions, particularly posterior distributions in Bayesian inference. Instead of sampling from the posterior (as in Markov Chain Monte Carlo methods), VI reformulates inference as an optimization problem.

Consider a model with unknown latent variables \(\mathbf{z}\), known variables \(\mathbf{x}\), and parameters \(\theta\) (in the case of a parametric model). We want to compute the posterior
\begin{equation}
    p_{\theta}(\mathbf{z}|\mathbf{x}) = \frac{p_{\theta}(\mathbf{x}|\mathbf{z}) p_{\theta}(\mathbf{z})}{p_{\theta}(\mathbf{x})},
\end{equation}
where we assumes the normalization constant \(p_{\theta}(\mathbf{x}) = \int p_{\theta}(\mathbf{x}, \mathbf{z}) d\mathbf{z}\) is intractable.

VI approximates \(p(\mathbf{z}|\mathbf{x})\) with a simpler distribution \(q_{\psi}(\mathbf{z})\) from a parametric family \(\mathcal{Q}\) (e.g., Gaussians) such that:
\begin{equation}
\begin{array}{l l}
    \psi^* &= \text{argmin}_{\psi} \; D_{\mathbb{KL}}(q_{\psi}(\mathbf{z}) \| p_{\theta}(\mathbf{z}|\mathbf{x})) \\
    & = \text{argmin}_{\psi} \mathbb{E}_{q_\psi(\mathbf{z})} \big[ q_{\psi}(\mathbf{z}) - \log \big( \frac{p_{\theta}(\mathbf{x}|\mathbf{z}) p_{\theta}(\mathbf{z})}{p_{\theta}(\mathbf{x})} \big) \big] \\
    & = \text{argmin}_{\psi} \underbrace{\mathbb{E}_{q_\psi(\mathbf{z})}[ \log q_\psi(\mathbf{z}) - \log p_{\theta}(\mathbf{x}|\mathbf{z})- \log p_\theta(\mathbf{z})]}_{\mathcal{L}(\theta, \psi|\mathbf{x})} + \log p_\theta(\mathbf{x}).
\end{array}
\end{equation}
\(\psi\) are known as the variational parameters, which we optimize to obtain our approximate distribution by minimizing \(\mathcal{L}(\theta, \psi | \mathbf{x}) = \mathbb{E}_{q_\psi(\mathbf{z})}[-\log p_\theta (\mathbf{x}, \mathbf{z}) + \log q_\psi(\mathbf{z})]\). This objective can be rewritten as maximizing the evidence lower bound (ELBO):
\begin{equation}
    L(\psi, \theta| \mathbf{x}) = \underbrace{\mathbb{E}_{q_\psi(\mathbf{z})}[\log p_\theta(\mathbf{x}|\mathbf{z})]}_{\text{expected log likelihood}} - \underbrace{D_{\mathbb{KL}}(q_\psi(\mathbf{z}) \| p_\theta(\mathbf{z}))}_{\text{KL from posterior to prior}}.
    \label{eq:elbo}
\end{equation}
The KL term acts as a regularization term, ensuring the approximate posterior does not diverge too much from the prior distribution. As the name implies, \(L(\theta, \psi | \mathbf{x})\) is a lower bound of the evidence \(\log p_{\theta}(\mathbf{x})\):
\begin{equation}
    L(\theta, \psi | \mathbf{x}) \leq \log p_{\theta}(\mathbf{x}).
\end{equation}

\section{Ablation Experiments}
\label{sec:ablation}

\subsection{Impact of Inducing Points}
\begin{figure}[h!]
    \centering
    \includegraphics[width=0.45\linewidth]{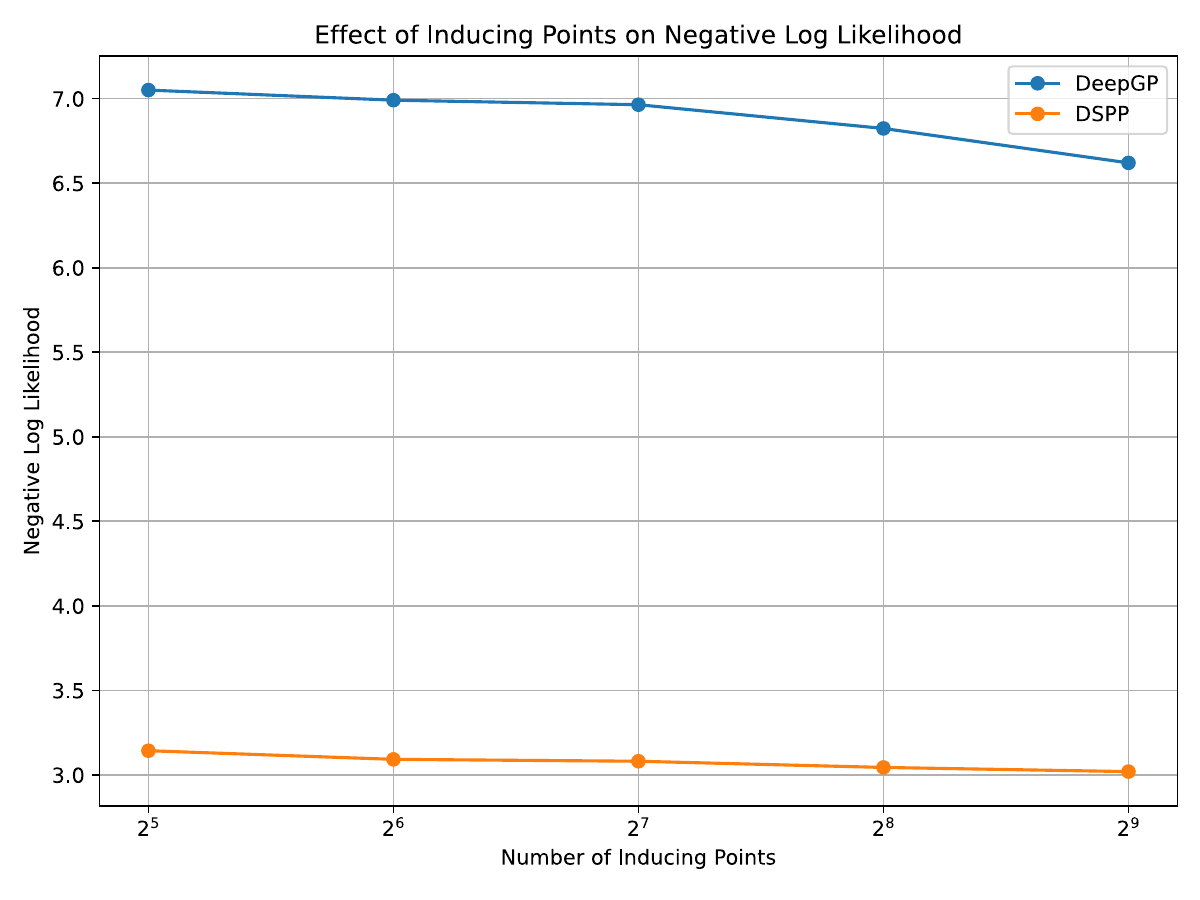}
    \includegraphics[width=0.45\linewidth]{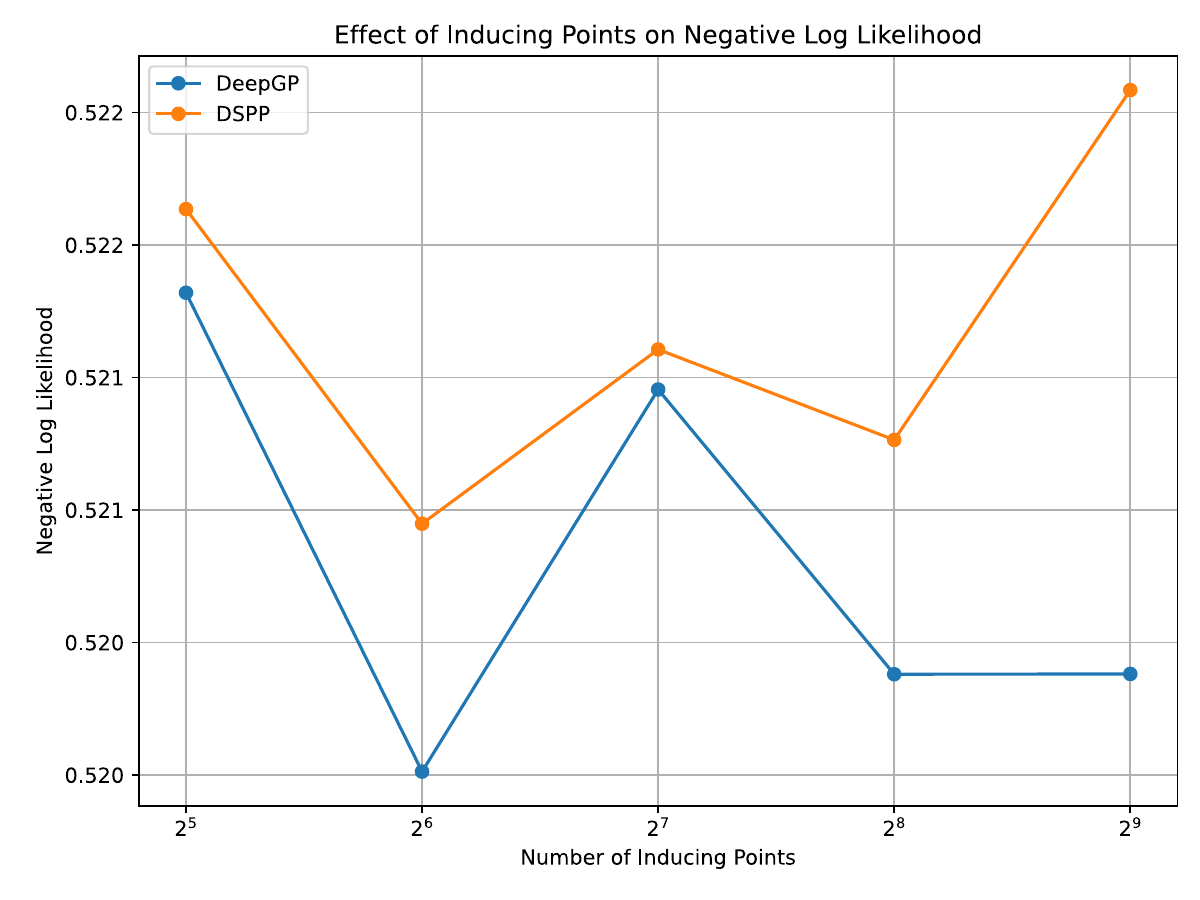}
    \caption{Negative log likelihood on test set against the number of inducing points. Left: CASP, Right: ESR}
    \label{fig:nllbruh}
\end{figure}

For this experiment, we were interested in the relationship between the number of inducing points and calibration error for our dataset. We considered DGPs/DSPPs with no hidden layers and train for 20 epochs with a learning rate of \(0.01\). We performed training and evaluation for \(M \in \{32, 64, 128, 256, 512\}\) inducing points. The results can be seen in Figure \ref{fig:nllbruh}.

\subsection{Impact of Depth/Number of Layers}
\begin{figure}[h!]
    \centering
    \includegraphics[width=0.45\linewidth]{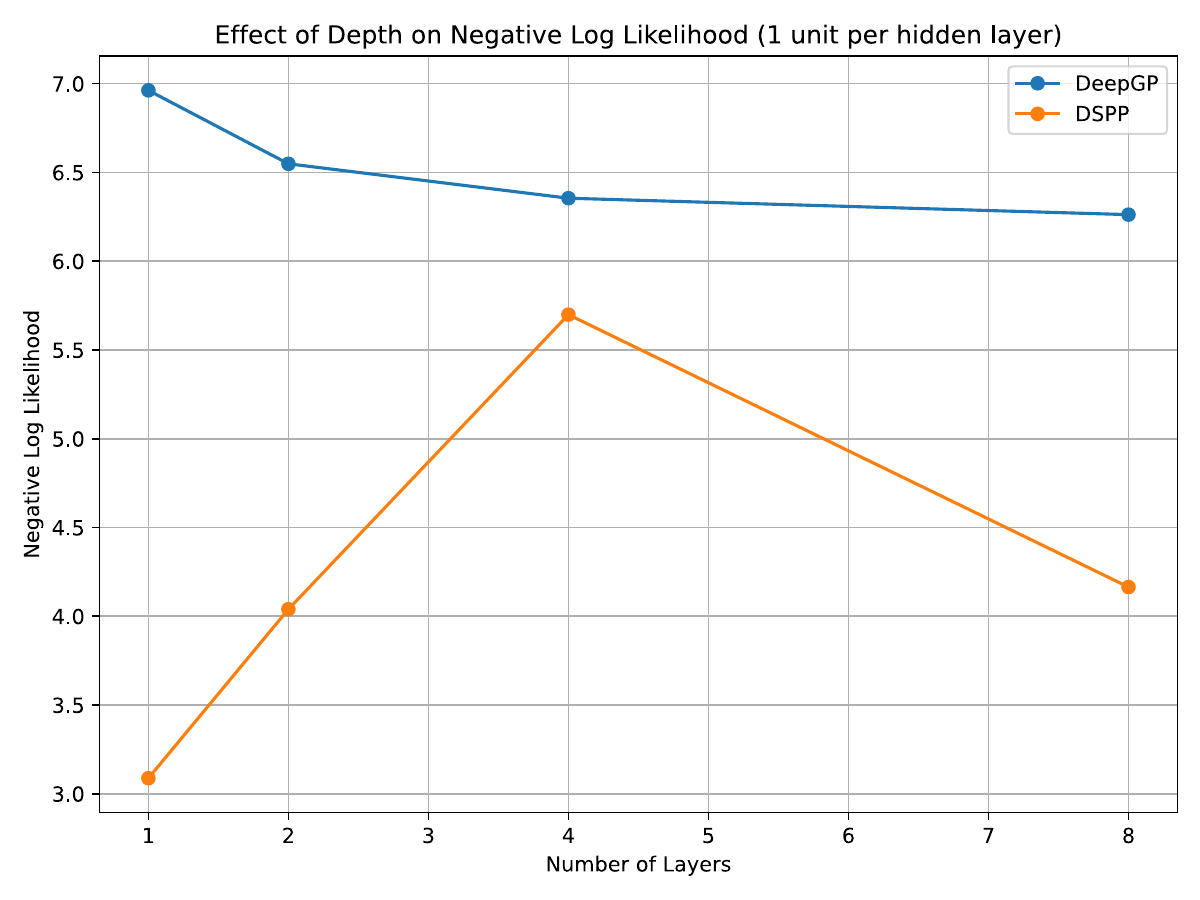}
    \includegraphics[width=0.45\linewidth]{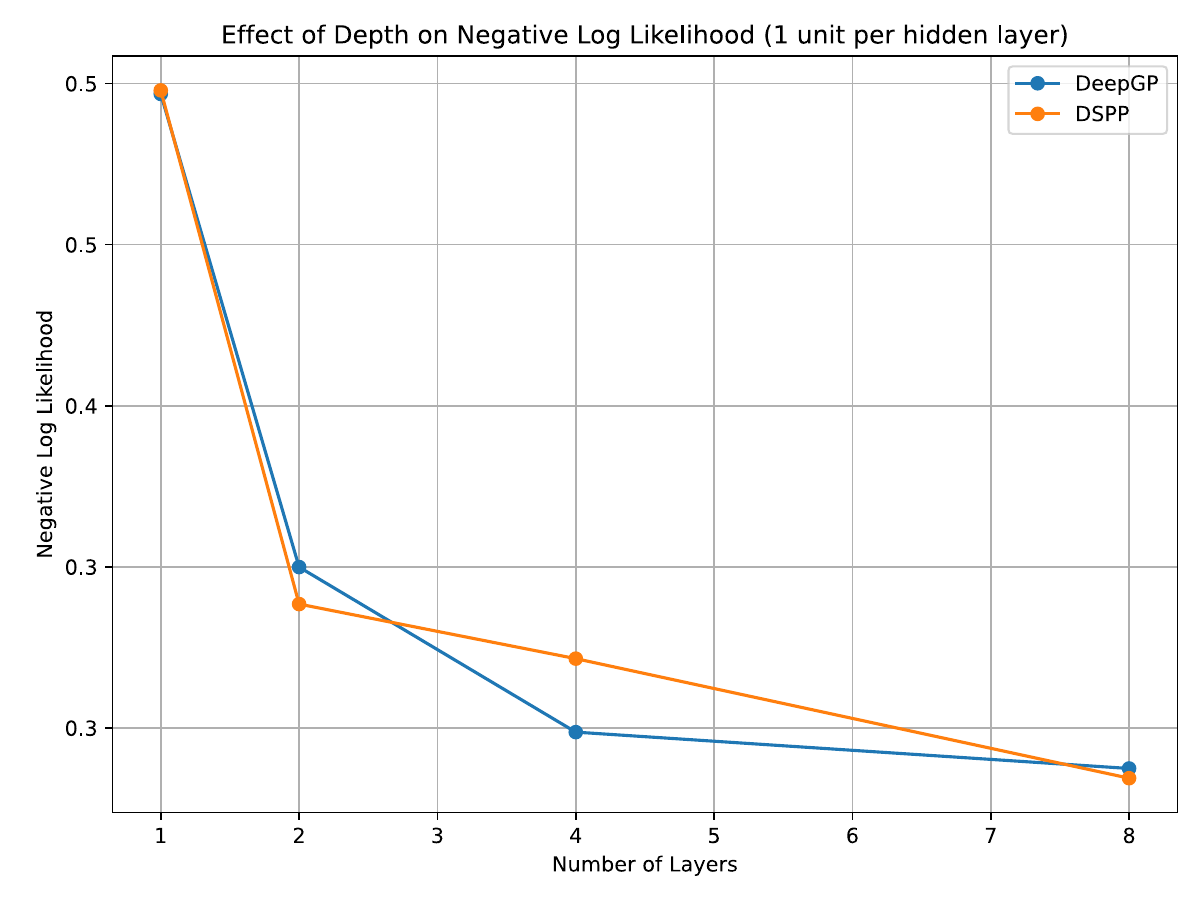}
    \caption{Negative log likelihood on test set against the number of hidden layers. Left: CASP, Right: ESR}
    \label{fig:nll}
\end{figure}

For this experiment, we were interested in the relationship between the number of and the negative log likelihood for our dataset. Both the DGP and DSPP were trained for 20 epochs with a learning rate of \(0.01\). For simplicity, we used a single GP per hidden layer with 128 inducing points and performed training and evaluation for \(d \in \{1,2,4,8\}\) number of layers. The results can be seen in Figure \ref{fig:nll}.

\section{Additional Distribution Shift Results}
\begin{figure}[h!]
    \centering
    \includegraphics[width=0.48\linewidth]{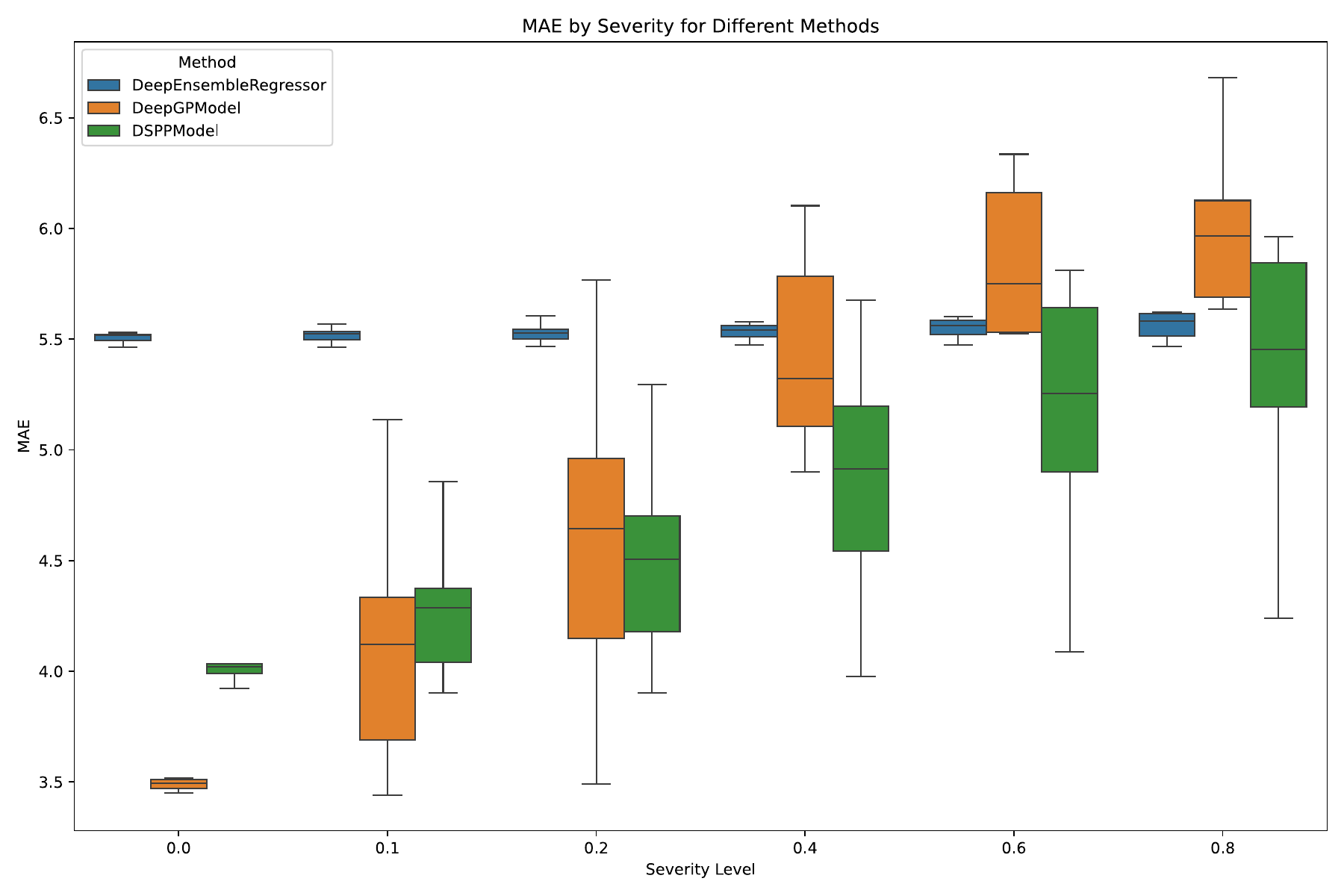}
    \includegraphics[width=0.48\linewidth]{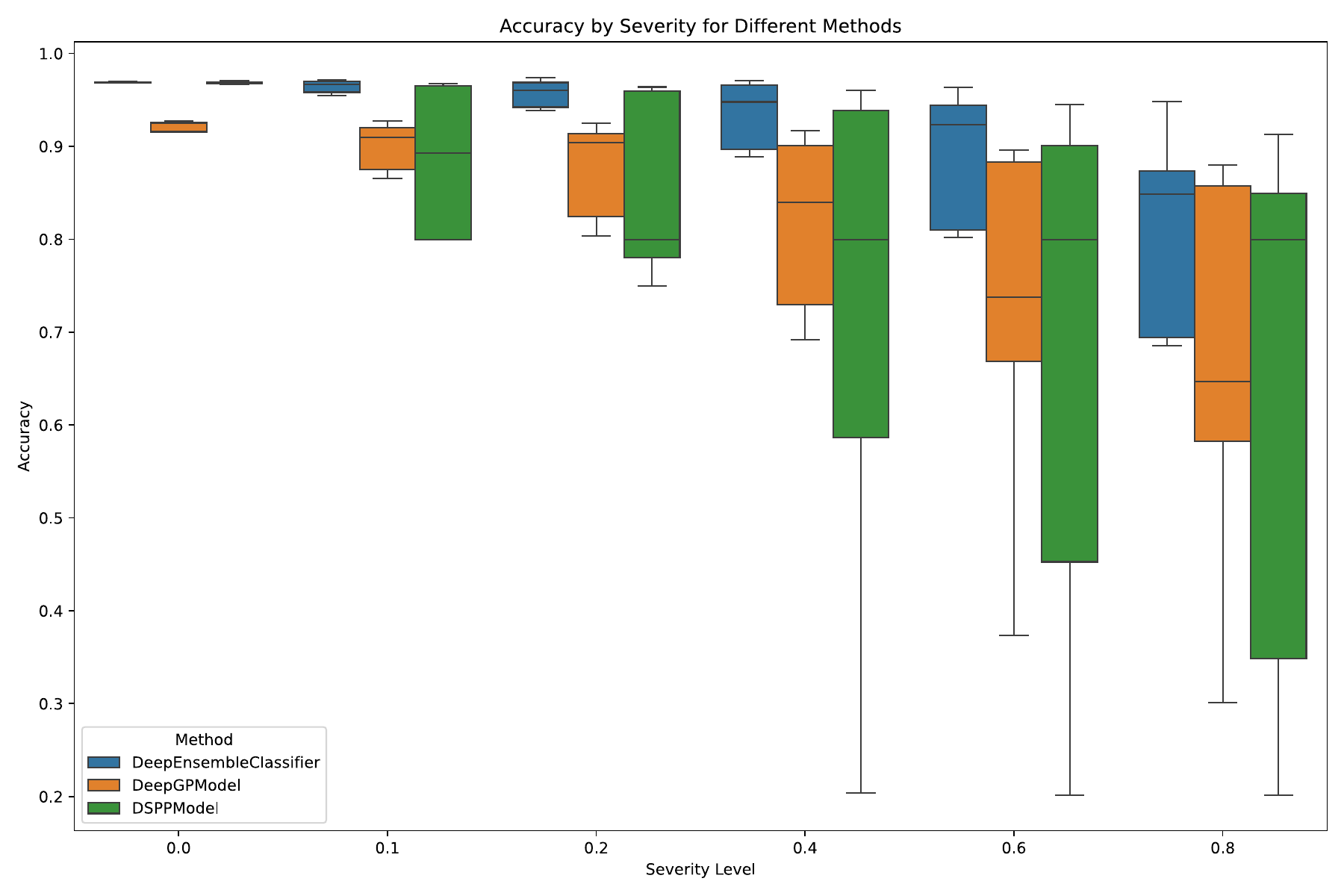}
    \caption{Box plots showing MAE and accuracy for the ESR classification (right) and CASP regression (left) tasks under distributional shift across three methods: Deep Ensemble (blue), DGP (orange), and DSPP (green).}
    \label{fig:boxplotMAE&ACC}
\end{figure}

Figure \ref{fig:boxplotMAE&ACC} (Left) presents the MAE under increasing shift severity for the CASP regression task. DGP initially achieves the lowest error but degrades more rapidly under shift. DSPP shows a more gradual increase in error and outperforms DGP at higher severities. The Deep Ensemble maintains stable but consistently higher error across all levels. Overall, DSPP demonstrates the best robustness and predictive accuracy under distributional shift.

In the ESR classification task, Figure \ref{fig:boxplotMAE&ACC} (Right) summarizes model accuracy across severity levels. The Deep Ensemble consistently maintains the highest accuracy across all shift intensities, showing strong robustness to corruption. While DSPP initially performs well, it experiences a sharper and less stable decline in accuracy as severity increases. In contrast, DGP exhibits a lower median accuracy at high shift levels but with tighter interquartile ranges, indicating more consistent performance. DSPP maintains a higher median accuracy than DGP at the most severe shifts, yet the wider spread in its predictions reflects greater variability and less reliability under strong distributional shifts.

\end{document}